\documentclass{article}
\usepackage{ijcai21}

\usepackage{times}
\usepackage{soul}
\usepackage{url}
\usepackage[utf8]{inputenc}
\usepackage[small]{caption}
\usepackage{graphicx}
\usepackage{amsmath}
\usepackage{amsthm}
\usepackage{booktabs}
\usepackage{algorithm}
\usepackage{algorithmic}
\usepackage{amssymb}
\usepackage{color}

\def\ie{\emph{i.e.}}
\def\eg{\emph{e.g.}}

\title{Structure Guided Lane Detection}

\author{
Jinming Su$^*$ \and
Chao Chen \and
Ke Zhang \and
Junfeng Luo$^*$ \and
Xiaoming Wei \And
Xiaolin Wei
\affiliations
Meituan
\emails
\{sujinming, chenchao60, zhangke21, luojunfeng, weixiaoming, weixiaolin02\}@meituan.com
}

\begin{document}

\maketitle

\begin{abstract}
Recently, lane detection has made great progress with the rapid development of deep neural networks and autonomous driving. 
However, there exist three mainly problems including characterizing lanes, modeling the structural relationship between scenes and lanes, and supporting more attributes (\eg, instance and type) of lanes.
In this paper, we propose a novel structure guided framework to solve these problems simultaneously. In the framework, we first introduce a new lane representation to characterize each instance. Then a top-down vanishing point guided anchoring mechanism is proposed to produce intensive anchors, which efficiently capture various lanes. 
Next, multi-level structural constraints are used to improve the perception of lanes. In the process, pixel-level perception with binary segmentation is introduced to promote features around anchors and restore  lane details from bottom up, a lane-level relation is put forward to model structures (\ie, parallel)  around lanes, and an image-level attention is used to adaptively attend different regions of the image from the perspective of scenes. With the help of structural guidance, anchors are effectively classified and regressed to obtain precise locations and shapes.
Extensive experiments on public benchmark datasets show that the proposed approach outperforms state-of-the-art methods with 117 FPS on a single GPU.
\end{abstract}
\let\thefootnote\relax\footnotetext{* Co-corresponding author.}

\section{Introduction}
Lane detection, which aims to detect lanes in road scenes, is a fundamental perception task and has a wide range of applications (\eg, ADAS~\cite{butakov2014personalized}, autonomous driving~\cite{chen2017end} and high-definition map production~\cite{homayounfar2019dagmapper}). Over the past years, lane detection has made significant progress and it is also used as an important element for tasks of road scene understanding, such as driving area detection~\cite{yu2020bdd100k}. 

\begin{figure}[t]
\centering
\includegraphics[width=1\linewidth]{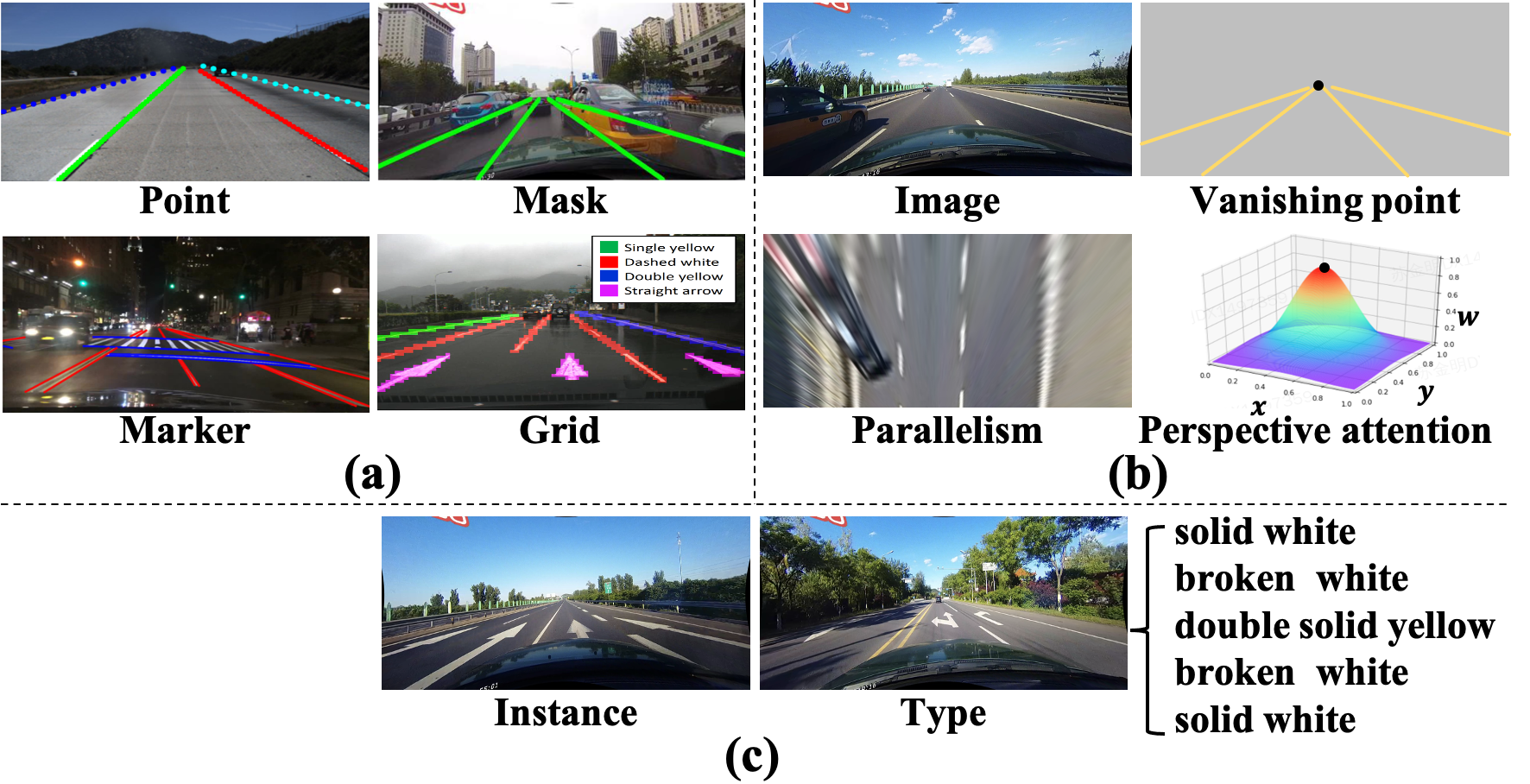}
\caption{Challenges of lane detection. (a) Various representation. There exist many kinds of annotations~\protect\cite{tusimple,pan2018spatial,yu2020bdd100k,lee2017vpgnet}, which makes it difficult to  characterize lanes in a unified way. (b) Underresearched scene structures. Lane location are strongly dependent on structural information, such as vanishing point (black point), parallelism in bird's eye view and distance attention caused by perspective. (c) More attributes to support. Lanes have more attributes such as instance and type, which should be predicted.}
\label{fig:motivation}
\end{figure}

To address the task of lane detection, lots of learning-based methods~\cite{pan2018spatial,qin2020ultra} have been proposed in recent years, achieving impressive performance on existing benchmarks~\cite{tusimple,pan2018spatial}. However, there still exist several challenges that hinder the development of lane detection. 
Frist, there lacks a unified and effective lane representation. As shown in (a) of Fig.~\ref{fig:motivation}, there exist various definitions including point~\cite{tusimple}, mask~\cite{pan2018spatial}, marker~\cite{yu2020bdd100k} and grid~\cite{lee2017vpgnet}, which are quite different in form for
different scenarios.
Second, it is difficult to model the structural relationship between scenes and lanes. As displayed in (b) of Fig.~\ref{fig:motivation}, the structural information depending on scenes, such as location of vanishing points and parallelism of lanes, is very useful, but there is no scheme to describe it.
Last, while predicting lanes, it is also important to predict other attributes including instance and type (see (c) of Fig.~\ref{fig:motivation}), but it is not easy to extend these for existing methods.
These three difficulties are especially difficult to deal with and greatly slow down the development of lane detection. Due to these difficulties, lane detection remains a challenging vision task.

\begin{figure*}[t]
\centering
\includegraphics[width=1\textwidth]{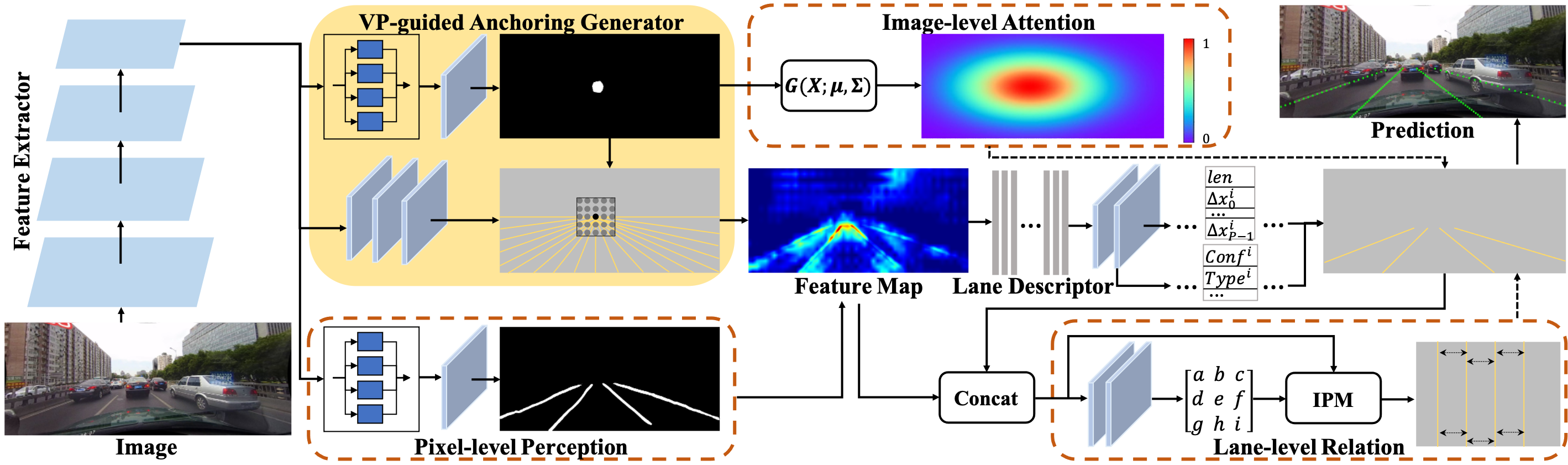}
\caption{Framework of our approach. We first extract the common features by the extractor, which provides features for vanishing point guided anchoring and pixel-level perception. The anchoring produces intensive anchors and perception utilizes binary segmentation to promote features around lanes. Promoted features are used to classify and regress anchors with the aid of lane-level relation and image-level attention. The dashed arrow indicates the supervision, and the supervision of vanishing point and lane segmentation is omitted in the figure.}
\label{fig:framework}
\end{figure*}

To deal with the first difficulty, many methods characterize lanes with simple fitted curves or masks. For examples, 
SCNN~\cite{pan2018spatial} treats the problem as a semantic segmentation task, and introduces slice-by-slice convolutions within feature maps, thus enabling message passing. For these methods, lanes are characterized as a special form (\eg, point, curve or mask), so it is difficult to support the format of marker or grid that usually has an uncertain number. Similarly, those who support the latter~\cite{lee2017vpgnet} do not support the former well.
To address the second problem, some methods use vanishing point or parallel relation as auxiliary information. For example, a vanishing point prediction task~\cite{lee2017vpgnet} is utilized to implicitly embed a geometric context recognition capability.
In these methods, they usually only pay attention to a certain kind of structural information or do not directly use it end-to-end, which leads to the structures not fully functioning and the algorithm complicated. 
For the last problem, some clustering- or detection-based methods are used to distinguish or classify instances. 
Line-CNN~\cite{li2019line} utilizes line proposals as references to locate traffic curves, which forces the method to learn the feature of lanes. 
To these methods, they can distinguish instances and even extend to more attributes, but they usually need extra computation and have many manually designed super-parameters, which leads to poor scalability.

Inspired by these observations and analysis, we propose a novel structure guided framework for lane detection, as shown in Fig.~\ref{fig:framework}. In order to characterize lanes, we propose a box-line based proposal method. In this method, the minimum circumscribed rectangle of the lane is used to distinguish instance, and its center line is used for structured positioning. 
For the sake of further improving lane detection by utilizing structural information, the vanishing point guided anchoring mechanism is proposed to generate intensive anchors (\ie, as few and accurate anchors as possible). In this mechanism, vanishing point is learned in a segmentation manner and used to produce structural anchors top-down, which can efficiently capture various lanes. Meanwhile, we put forward multi-level structure constraints to improve the perception of lanes. In the process, the pixel-level perception is used to improve lane details with the help of lane binary segmentation, the lane-level relation aims at modeling the parallelism properties of inter-lanes by Inverse Perspective Mapping (IPM) via a neural network, and image-level attention is to attend the image with adaptive weights from the perspective of scenes. Finally, features of lane anchors under structural guidance are extracted for accurate classification, regression and the prediction of other attributes. 
Experimental results on CULane and Tusimple datasets verify the effectiveness of the proposed method which achieves state-of-the-art performance and run efficiently at 117 FPS.


The main contributions of this paper include: 1) we propose a structure guided framework for lane detection, which characterize lanes and can accurately class, locate and restore the shape of unlimited lanes. 2) we introduce a vanishing point guided anchoring mechanism, in which the vanishing point is predicted and used to produce intensive anchors, which can precisely capture lanes. 3) we put forward the multi-level structural constraints, which are used to sense pixel-level unary details, model lane-level pair-wise relation and adaptively attend image-level global information.

\section{Related Work}
In this section, we review the related works that aim to resolve the challenges of lane detection in two aspects.

\subsection{Traditional Methods}
To solve the problem of lane detection, traditional methods are usually based on hand-crafted features 
by detecting shapes of markings and fitting 
the spline. 
~\cite{veit2008evaluation} presents a comprehensive overview of features used to detect road markings. And~\cite{wu2012practical} uses Maximally Stable Extremal Regions features and performs the template matching to detect multiple road markings. However, there approaches often fail in unfamiliar conditions.

\subsection{Deep Learning based Methods}
With the development of deep learning, methods~\cite{pizzati2019enhanced,van2019end,guo2020gen} based on deep neural networks achieve progress in lane detection. SCNN~\cite{pan2018spatial} generalizes traditional deep layer-by-layer convolutions to enable message passing between pixels across rows and columns.
ENet-SAD~\cite{hou2019learning} presents a knowledge distillation approach, 
which allows a model to learn from itself 
without any additional supervision or labels. PolyLaneNet~\cite{tabelini2020polylanenet} adopts a polynomial representation for the lane markings,
and outputs polynomials via the deep polynomial regression. UltraFast~\cite{qin2020ultra} treats the process of lane detection as a row-based selecting problem using global features. CurveLanes~\cite{xu2020curvelane} proposes a lane-sensitive architecture search framework to automatically capture both long-ranged coherent and accurate short-range curve information. 

In these methods, different lane representations are adopted and some structural information is considered for performance improvement. However, these methods are usually based on the powerful learning ability of neural networks to learn the fitting or shapes of lanes, and the role of scene-related structural information for lanes has not been paid enough attention to and discussed.

\section{The Proposed Approach}
To address these difficulties (i.e., characterizing lanes, modeling the relationship between scenes and lanes, and supporting more attributes), we propose a novel structure guided framework for lane detection, denoted as \textbf{SGNet}. In this framework, we first introduce a new lane representation. Then a top-down vanishing point guided anchoring mechanism is proposed, and next multi-level structure constraints is used. Details of the proposed approach are described as follows.

\subsection{Representation}
For adapting to different styles of lane annotation, we introduce a new box-line based method for lane representation. Firstly, we calculate the minimum circumscribed rectangle $R$ (``box'') with the height $h$ and width $w$ for the lane instance $L_{lane}$. For this rectangle, center line $L_{center}$ (``line'') perpendicular to the short side is obtained. And the angle between the positive $X$-axis and $L_{center}$ in clockwise direction is $\theta$. In this manner, $L_{center}$ provides the position of the lane instance, and $h$ and $w$ restrict the areas involved. Based on $R$ and $L_{center}$ , lane prediction based on points, masks, markers, grids and other formats can be performed. In this paper, the solution based on key points of lane detection is taken just because of the point-based styles of lane annotation in public datasets (\eg, CULane~\cite{tusimple} and Tusimple~\cite{pan2018spatial}).

Inspired by existing methods~\cite{li2019line,chen2019pointlanenet,qin2020ultra}, we define key points of the lane instance with equally spaced $y$ coordinates $Y=\{y_i\}$ and $y_i = \frac{H}{P - 1} \cdot i (i=1,2,...,P-1)$, where $P$ means the number of all key points through image height, which is fixed on images with same height $H$ and width $W$. Accordingly, the $x$ coordinates of the lane is expressed as $X = \{x_i\}$. For the convenience of expression, the straight line equation of $L_{center}$ is defined as 
\begin{equation}\label{eq:straight_line}
ax + by + c = 0, a\neq 0 \ or \ b \neq 0 
\end{equation}
where $a$, $b$ and $c$ can be easily computed by $\theta$ and any point on $L_{center}$. Next, when the $y$ coordinate of the center line is $y_i$, we can compute the corresponding $x$ coordinate as
\begin{equation}\label{eq:straight_line}
x_i = L_{center}(y_i) = \frac{-c-by_i}{a}, a \neq 0.
\end{equation}
Then, we define the offset of $x$ coordinate $\Delta X$   between the lane $L_{lane}$ and center line $L_{center}$ as
\begin{equation}\label{eq:Delta_x}
\begin{aligned}
\Delta X &=  \{\Delta x_i\} = \{ x_i - \frac{-c-by_i}{a} \},  \\
X &= \{\frac{-c-by_i}{a} \} + \Delta X.
\end{aligned}
\end{equation}
Therefore, based on $L_{center}$ and $\Delta X$, we can calculate the lane instance $L_{lane}$. Usually, it is easier to learn $L_{center}$ and $\Delta X$ than the directly fitting key points of $L_{lane}$.

\begin{figure}[t]
\centering
\includegraphics[width=1\linewidth]{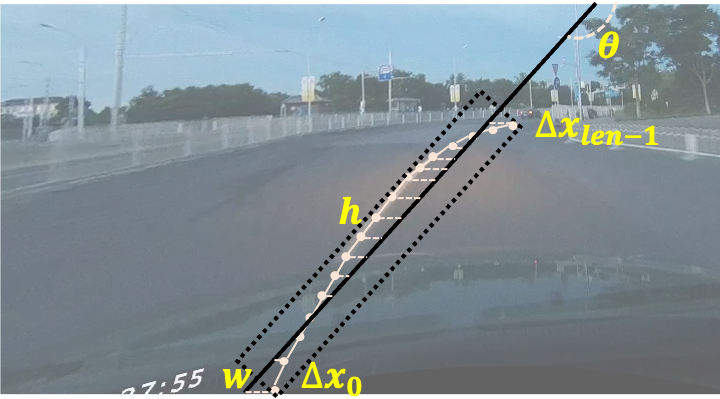}
\caption{Lane representation.}
\label{fig:lane_representation}
\end{figure}

\subsection{Feature Extractor}
To see Fig.~\ref{fig:framework}, SGNet takes ResNet~\cite{he2016deep} as the feature extractor, which is modified to remove the last global pooling and fully connected layers for the pixel-level
prediction task. Feature extractor has five residual modules for encoding, named as $\mathcal{E}_i(\pi_i)$ with parameters $\pi_i (i=1,2,...,5)$. To obtain larger feature maps, we convolve $\mathcal{E}_5(\pi_5)$ by a convolutional layer with 256 kernels of $3 \times 3$ and then $\times2$ upsample the features, followed by an element-wise summation with $\mathcal{E}_4(\pi_4)$ to obtain $\mathcal{E}_4'(\pi_4')$. Finally, for a $H \times W$ input image, a $\frac{H}{16} \times \frac{W}{16}$ feature map is output by the feature extractor.

\subsection{Vanishing Point Guided Anchoring}
In order to learn the lane representation, there are two main ways to learn the center line $L_{center}$ and $x$ offset $\Delta X$. The first way is to learn the determined $L_{center}$ directly with angle, number and position regression, which is usually difficult to achieve precise results because of the inherent difficulty of regression tasks. The second way is based on mature detection tasks, using dense anchors to classify,  regress and then obtain proposals representing the lane instance. And the second one has been proved to work well in general object detection tasks, so we choose it as our base model. 

To learn the center line $L_{center}$ and $x$ offset $\Delta X$ well, we propose a novel vanishing point guided anchoring mechanism (named as VPG-Anchoring). The vanishing point (VP) provides strong characterization of geometric scene, representing the end of the road and also the ``virtual'' point where the lanes intersect in the distance. Since VP is the intersection point of lanes, lanes in the scene must pass through VPs, and lines that do not pass through VPs are not lanes in the scene with high probability. Therefore, dense lines radiated from VPs can theoretically cover all lanes in the image, which is equivalent to reducing the generation space of anchors from $\mathbb{R}^{H \times W \times N_{proposal}}$ to $\mathbb{R}^{N_{proposal}}$. $N_{proposal}$ represents the number of anchors generated at one pixel.

As shown in Fig.~\ref{fig:framework}, the features map $\mathcal{E}'_4(\pi'_4)$ is feed to VPG-Anchoring. In the mechanism, VP is predicted by a simple branch, which is implemented by a multi-scale context-aware atrous spatial pyramid pooling (ASPP)~\cite{chen2018encoder} followed by a convolutional layer with 256 kernels of $3 \times 3$ and a softmax activation. The VP prediction branch is denoted as $\phi_{\mathcal{V}}({\pi_\mathcal{V}})$ with parameters $\pi_\mathcal{V}$.

Usually, VP is not annotated in lane datasets, such as CULane~\cite{pan2018spatial}, so we average the intersection points of the center lines of all lane instances and get the approximate VP. In addition, a single point is usually difficult to predict, so we expand the area of VP to a radius of 16 pixels and use segmentation algorithm to predict. To achieve this, we expect the output of $\phi_{\mathcal{V}}({\pi_\mathcal{V}})$ to approximate the ground-truth masks of VP (represented as $G_{\mathcal{V}}$) by minimizing the loss
\begin{equation}\label{eq:vp_loss}
\begin{aligned}
{\mathcal{L}}_{\mathcal{V}} = BCE(\phi_{\mathcal{V}}({\pi_\mathcal{V}}), G_{\mathcal{V}}),
\end{aligned}
\end{equation}
where $BCE(\cdot,\cdot)$ represents the pixel-level binary cross-entropy loss function.

\begin{figure}[t]
\centering
\includegraphics[width=1\linewidth]{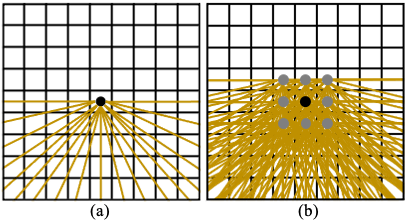}
\caption{VP-guided anchoring mechanism. Anchors (golden lines) generated based on (a) the vanishing point (black point) and (b) the area around vanishing point (black and gray points). }
\label{fig:anchor}
\end{figure}

In order to ensure that generated anchors are dense enough, we choose a $W_{anchor} \times W_{anchor}$ rectangular area with VP as the center, and take one point every $S_{anchor}$ to generate anchors. For each point, anchors are generated every $A_{anchor}$ angle ($A_{anchor} \in [0, 180]$) as shown in Fig.~\ref{fig:anchor}. 
In this way, anchors are targeted, intensive and not redundant, compared with general full-scale uniform generation and even specially designed methods for lanes~\cite{li2019line}. 
Note that anchors run through the whole image, and only the part below VP is shown for convenient display in Figs.~\ref{fig:framework} and~\ref{fig:anchor}.

\subsection{Classification and Regression}
In order to classify and regress the generated anchors, we extract high-level feature maps based on  $\mathcal{E}_4(\pi_4)$ with several convolutional layers. The feature map is named as $\text{F}_{\mathcal{A}} \in \mathbb{R}^{H' \times W' \times C'}$, where $H', W'$ and $C'$ are the height, width and channel of $\text{F}_{\mathcal{A}}$. For each anchor $L_{lane}$, the channel-level features of each point on anchors are extracted from $\text{F}_{\mathcal{A}}$ to obtain lane descriptor $\text{D}_{\mathcal{A}} \in  \mathbb{R}^{H' \times C'}$, which are used to classify the existence $Conf^{L_{lane}}$ and regress $x$ offsets $\Delta X^{L_{lane}}$ including the length $len$ of lanes. To learn these, we expect the output to approximate the ground-truth existence $GConf^{L_{lane}}$ and $x$ offsets $G\Delta X^{L_{lane}}$  by minimizing the loss
\begin{equation}
\begin{aligned}
{\mathcal{L}}_{\mathcal{C}} &= \sum_{L_{lane} = 0}^{L-1}BCE(Conf^{L_{lane}}, GConf^{L_{lane}}),  \\
{\mathcal{L}}_{\mathcal{R}} &= \sum_{L_{lane} = 0}^{L-1}SL1(\Delta X^{L_{lane}}, G\Delta X^{L_{lane}}),
\end{aligned}
\label{eq:class_regress_loss_ori}
\end{equation}
where $SL1(\cdot, \cdot)$ means smooth L1 loss and L means the number of proposals. Finally, Line-NMS~\cite{li2019line} is used to obtain the finally result with confidence thresholds.

\subsection{Multi-level Structure Constraints}
In order to further improve lane perception, we ask for the structural relationship between scenes and lanes, and deeply explore the pixel-level, lane-level and image-level structures.

\paragraph{Pixel-level Perception.}
The top-down VPG-Anchoring mechanism covers the structures and distribution of lanes. At the same time, there is a demand of bottom-up detail perception, which ensures that lane details are restored and described more accurately. For the sake of improving the detail perception, we introduce lane segmentation branch to location lane locations and promote pixel-level unary details. As shown in Fig.~\ref{fig:framework}, the lane segmentation branch has the same input and similar network structure with the VP prediction branch. The lane segmentation branch is denoted as $\phi_{\mathcal{P}}({\pi_\mathcal{P}})$ with parameters $\pi_\mathcal{P}$. To segment lanes, we expect the output of $\text{P}_{\mathcal{P}} = \phi_{\mathcal{P}}({\pi_\mathcal{P}})$ to approximate the ground-truth masks of binary lane mask (represented as $G_{\mathcal{P}}$) by minimizing the loss
\begin{equation}\label{eq:pixel-level_loss}
\begin{aligned}
{\mathcal{L}}_{\mathcal{P}} = BCE(\text{P}_{\mathcal{P}}, G_{\mathcal{P}}).
\end{aligned}
\end{equation}
To promote the pixel-level unary details, we weight the input features $\text{F}_{\mathcal{A}}$ by the following operation
\begin{equation}\label{eq:Delta_x}
\begin{aligned}
\text{M}_{\mathcal{A}} = \text{F}_{\mathcal{A}} \otimes \text{P}_{\mathcal{P}} + \text{F}_{\mathcal{A}},
\end{aligned}
\end{equation}
where $M_{\mathcal{A}}$ are feed to classify and regress instead of $\text{F}_{\mathcal{A}}$.

\begin{figure*}[t]
\centering
\includegraphics[width=1\textwidth]{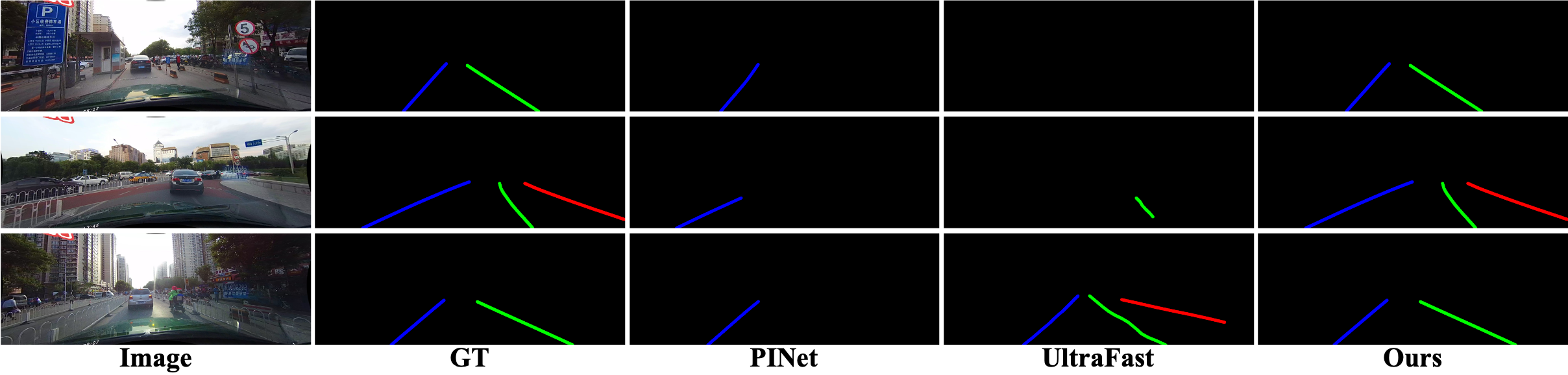}
\caption{Qualitative comparisons of the state-of-the-art algorithms and our approach.}
\label{fig:result}
\end{figure*}

\begin{table*}[t]
\centering
\renewcommand\arraystretch{1.0}
\small
\begin{tabular}{c | c | c c c c c c c c c |c}
\hline
 & \textbf{Total}
 & \textbf{Normal}
 & \textbf{Crowd}
 & \textbf{Dazzle}
 & \textbf{Shadow}
 & \textbf{No line}
 & \textbf{Arrow}
 & \textbf{Curve}
 & \textbf{Cross}
 & \textbf{Night}
  & \textbf{FPS}
 \\
\hline
DeepLabV2-50
& 66.70& 87.40 & 64.10 & 54.10 & 60.70 & 38.10 & 79.00 & 59.80 & 2505 & 60.60 &  - \\
SCNN
& 71.60 & 90.60 & 69.70 & 58.50 & 66.90 & 43.40 & 84.10 & 64.40 & 1990 & 66.10 &  8\\
FD
& -      & 85.90 & 63.60 & 57.00 & 59.90 & 40.60 & 79.40 & 65.20 & 7013 & 57.80 & - \\
ENet-SAD
& 70.80 & 90.10 & 68.80 & 60.20 & 65.90 & 41.60 & 84.00 & 65.70 & 1998 & 66.00 &  75 \\
PointLane
&70.20 & 88.00 & 68.10 & 61.50 & 63.30 & 44.00 & 80.90 & 65.20 &  1640 & 63.20 & - \\
RONELD
 & 72.90 & -     & - & - & - & - & - & - & - & - & - \\ 
PINet
& 74.40 & 90.30 & 72.30 & 66.30 & 68.40 & {\color{blue}{\textbf{49.80$_{3}$}}} & 83.70 & 65.60 & {\color{blue}{\textbf{1427$_{3}$}}} & 67.70 & 25 \\
ERFNet-E2E
& 74.00 & {\color{blue}{\textbf{91.00$_{3}$}}} & {\color{blue}{\textbf{73.10$_{3}$}}} & 64.50 & {\color{green}{\textbf{74.10$_{2}$}}} & 46.60 & {\color{blue}{\textbf{85.80$_{3}$}}} & {\color{red}{\textbf{71.90$_{1}$}}} 
																				    & 2022 & 67.90 & - \\
IntRA-KD
& 72.40 & - &- &- &- &- &- &- &-&-& 98 \\
UltraFast-18
& 68.40 & 87.70 & 66.00 & 58.40 & 62.80 & 40.20 & 81.00 & 57.90 & 1743 & 62.10 & {\color{red}{\textbf{323$_{1}$}}}\\ 
UltraFast-34
& 72.30 & 90.70 & 70.20 & 59.50 & 69.30 & 44.40 & 85.70 & {\color{blue}{\textbf{69.50$_{3}$}}} & 2037 & 66.70 & {\color{green}{\textbf{175$_{2}$}}}\\ 
CurveLanes
& {\color{blue}{\textbf{74.80$_{3}$}}} & 90.70 & 72.30 & {\color{green}{\textbf{67.70$_{2}$}}} & 70.10 & 49.40 & {\color{blue}{\textbf{85.80$_{3}$}}} & 68.40 & 1746 & {\color{blue}{\textbf{68.90$_{3}$}}} &  - \\
\hline
\textbf{Ours-Res18}                   & {\color{green}{\textbf{76.12$_{2}$}}} & {\color{green}{\textbf{91.42$_{2}$}}} & {\color{green}{\textbf{74.05$_{2}$}}} & {\color{blue}{\textbf{66.89$_{3}$}}} & {\color{blue}{\textbf{72.17$_{3}$}}} & {\color{green}{\textbf{50.16$_{2}$}}} & {\color{green}{\textbf{87.13$_{2}$}}} & 67.02 & {\color{red}{\textbf{1164$_{1}$}}} & {\color{green}{\textbf{70.67$_{2}$}}} & {\color{blue}{\textbf{117$_{3}$}}}\\
\textbf{Ours-Res34}                   & {\color{red}{\textbf{77.27$_{1}$}}} & {\color{red}{\textbf{92.07$_{1}$}}} & {\color{red}{\textbf{75.41$_{1}$}}} & {\color{red}{\textbf{67.75$_{1}$}}} 
						    	& {\color{red}{\textbf{74.31$_{1}$}}} & {\color{red}{\textbf{50.90$_{1}$}}} & {\color{red}{\textbf{87.97$_{1}$}}} & {\color{green}{\textbf{69.65$_{2}$}}} 
							& {\color{green}{\textbf{1373$_{2}$}}} & {\color{red}{\textbf{72.69$_{1}$}}} & 92\\
\hline
\end{tabular}
\caption{Comparisons with state-of-the-art methods on CULane dataset. F1-measure score (``\%'' is omitted) is used to evaluate the results of total and 8 sub-categories. For Cross, only FP are shown. The top three results are in {\color{red}{\textbf{red$_{1}$}}}, {\color{green}{\textbf{green$_{2}$}}} and {\color{blue}{\textbf{blue$_{3}$}}} fonts with a footnote.}
\label{tab:performance_on_culane}
\end{table*}

\paragraph{Lane-level Relation.}
In fact, lanes conform to certain rules in the construction process, and the most important one is that the lanes are parallel. Due to imaging reasons, this relationship is no longer maintained after perspective transformation, but it can be modeled potentially. To model the lane-level relation, we conduct IPM by the $H$ Matrix~\cite{neven2018towards} via a neural network. After learning $H$, the lane instance $L_{lane}$ can be transformed to $L'_{lane}$ on bird's eye view, where different instances are parallel. 
Formally, we define the relationship between lanes as follows. For two lane instances $L_{lane1}$ and $L_{lane2}$ in the image, they are projected to the bird's-eye view through the learned $H$ matrix, and the corresponding instance $L'_{lane1}$ and $L'_{lane2}$ are obtained. The two instances can be fitted to the following linear equations:
\begin{equation}\label{eq:Delta_x}
\begin{aligned}
a_1 * x + b_1 * y + c_1 &= 0, \\
a_2 * x + b_2 * y + c_2 &= 0.
\end{aligned}
\end{equation}
In these two equations, under the condition that y is equal, the difference of x is always constant. Thus we can get that $a_1 * b_2 = a_2 * b_1$. Expanding to all instances, lane-level relation can be formulated as 
\begin{equation}\label{eq:lane-level_loss}
\begin{aligned}
L_{\mathcal{L}} = \sum_{i = 0, j = 0, i \neq j}^{L - 1} L1(a_i b_j - a_j b_i).
\end{aligned}
\end{equation}

\paragraph{Image-level Attention.}
In the process of camera imaging, distant objects are small after projection. Usually, the distant information of lanes is not prominent visually, but they are equally important. After analysis, it is found that the distance between lanes and VP reflects the inverse proportion to scales in imaging. Therefore, we generate perspective attention map $\text{PAM}$ based on VP, which is based on the strong assumption that the attention and distance after imaging satisfies two-dimensional gaussian distribution. PAM ensures the attention of different regions by adaptively restricting the classification and regression loss (from Eq.~\ref{eq:class_regress_loss_ori}) as follows.
\begin{equation}\label{eq:class_regress_loss_after}
\begin{aligned}
L_{\mathcal{I}} =& \sum_{L_{lane} = 0}^{L-1} \sum_{p = 0}^{P - 1} L1(\Delta x^{L_{lane}}_p, G\Delta x^{L_{lane}}_p) \\
 & \cdot (1 +  |E(x^{L_{lane}}_p, y^{L_{lane}}_p)|), 
\end{aligned}
\end{equation}
where $|\cdot|$ means normalized to [0, 1].

By taking the losses of Eqs.(\ref{eq:vp_loss}),(\ref{eq:class_regress_loss_ori}),(\ref{eq:pixel-level_loss}),(\ref{eq:lane-level_loss}) and (\ref{eq:class_regress_loss_after}), the overall learning objective can be formulated as follows:
\begin{equation}\label{eq:total_loss}
\begin{aligned}
\min _{\mathbb{P}} \mathcal{L}_\mathcal{V} + \mathcal{L}_\mathcal{C} + \mathcal{L}_\mathcal{R} + \mathcal{L}_\mathcal{P} + \mathcal{L}_\mathcal{L} + \mathcal{L}_\mathcal{I},
\end{aligned}
\end{equation}
where $\mathbb{P}$ is the set of $\{\{\pi_i\}^5_{i=1}, \pi'_4, \pi_\mathcal{V}, \pi_\mathcal{C}, \pi_\mathcal{R}, \pi_\mathcal{P}, \pi_\mathcal{L}\}$, and $\pi_\mathcal{C}, \pi_\mathcal{R}$ and $\pi_\mathcal{L}$ are the parameters of classification, regression and lane-level relation subnetworks, respectively.

\section{Experiments and Results}
\subsection{Experimental Setup}

\paragraph{Dataset.}
To evaluate the performance of the proposed method, we conduct experiments on CULane~\cite{pan2018spatial} and Tusimple~\cite{tusimple} dataset. CULane dataset has a split with 88,880/9,675/34,680 images for train/val/test and Tusimple dataset is divided into three parts: 3,268/358/2,782 for train/val/test.

\paragraph{Metrics.}
For CULane, we use F1-measure score as the evaluation metric. Following~\cite{pan2018spatial}, we treat each lane as a line with 30 pixel width and compute the intersection-over-union (IoU) between groundtruths and predictions with a threshold of 0.5 to 
For Tusimple, the official metric (Accuracy) is used as the evaluation criterion, which evaluates the correction of predicted lane points.

\paragraph{Training and Inference.}
We use Adam optimization algorithm to train our network end-to-end by optimizing the loss in Eq. (\ref{eq:total_loss}). In the optimization process, the parameters of feature extractor are initialized by the pre-trained ResNet-18/34 model and ``poly'' learning rate policy are employed for all experiments. The training images are resized to the resolution of $360 \times 640$ for faster training, and applied affine and flipping. And we train the model for 10 epochs on CULane and 60 epochs on TuSimple.  Moreover, we empirically and experimentally set the number of points $P=72$, the width of rectangular $W_{anchor} = 40$, anchor strides $S_{anchor} = 5$ and anchor angle interval $A_{anchor} = 5$.

\begin{table}[t]
\centering
\renewcommand\arraystretch{1.0}
\small
\begin{tabular}{c | c | c }
\hline
 & \textbf{Accuracy}  & \textbf{FPS}   \\
\hline
DeepLabV2-18
& 92.69 & 40 \\
DeepLabV2-34
& 92.84 & 20 \\
SCNN
& {\color{green}{\textbf{96.53$_{2}$}}} & 8 \\
FD
& 94.90 & - \\
ENet-SAD
& {\color{red}{\textbf{96.64$_{1}$}}} & {\color{blue}{\textbf{75$_{3}$}}} \\
Cascaded-CNN
& 95.24 & 60 \\
PolyLaneNet
& 93.36 & {\color{red}{\textbf{115$_{1}$}}} \\
\hline
\textbf{Ours-Res34} & {\color{blue}{\textbf{95.87$_{3}$}}} & {\color{green}{\textbf{92$_{2}$}}} \\
\hline
\end{tabular}
\caption{Comparisons with state-of-the-arts on Tusimple.}
\label{tab:performance_on_tusimple}
\end{table}

\subsection{Comparisons with State-of-the-art Methods}
We compare our approach with state-of-the-arts including DeeplabV2~\cite{chen2017deeplab}, SCNN~\cite{pan2018spatial}, FD~\cite{philion2019fastdraw}, ENet-SAD~\cite{hou2019learning} , PointLane~\cite{chen2019pointlanenet}, RONELD~\cite{chng2020roneld}, PINet~\cite{ko2020key}, ERFNet-E2E~\cite{yoo2020end}, IntRA-KD~\cite{hou2020inter}, UltraFast~\cite{qin2020ultra}, CurveLanes~\cite{xu2020curvelane}, Cascaded-CNN~\cite{pizzati2019lane} and PolyLaneNet~\cite{tabelini2020polylanenet}.

We compare our approach with 10 state-of-the-art methods on CULane dataset, as listed in Tab.~\ref{tab:performance_on_culane}. Comparing our ResNet34-based method with others, we can see that the proposed method consistently outperforms other methods across total and almost all categories. For the total dataset, our method is noticeably improved from 74.80\% to 77.27\% compared with the second best method. Also, it is worth noting that our method is significantly better on Crowd (+2.31\%), Arrow (+2.17\%) and Night (+3.79\%) compared with second best methods, respectively. In addition, we also obviously lower FP on Cross by 3.78\% relative to the second best one. As for Curve, we are slightly below the best method (ERFNet-E2E), which conducts special treatment for curve points while maybe damaging other categories. Moreover, our method has a faster FPS than almost all results. These observations present the efficiency and robustness of our proposed method and validate that VPG-Anchoring and multi-level structures are useful for the task of lane detection.

Some examples generated by our approach and other state-of-the-art algorithms are shown in Fig.~\ref{fig:result}. We can see that lanes can be detected with accurate location and precise shape by the proposed
method, even in complex situations. These visualizations indicate that the proposed lane representation has a good characterization of lanes, and also show the superiority of the proposed method.

Moreover, we list the comparisons on Tusimple as shown in Tab.~\ref{tab:performance_on_tusimple}. It can be seen that our method is competitive in highway scenes without adjustment, which further proves the effectiveness of structural information for lane detection.

\subsection{Ablation Analysis}
To validate the effectiveness of different components of the proposed method, we conduct several experiments on CULane to compare the performance variations of our methods.

\begin{table}[t]
\centering
\renewcommand\arraystretch{1.0}
\small
\begin{tabular}{c| c c c c | c}
\hline
& \textbf{VPG-A} & \textbf{Pixel} & \textbf{Lane} & \textbf{Image} & \textbf{Total} \\
\hline
Base & & & & &  71.98 \\
Base+V-F &\checkmark & & & & 74.08     \\
Base+V &\checkmark & & & & 74.27   \\
Base+V+P &\checkmark & \checkmark & & & 76.30   \\
Base+V+P+L &\checkmark & \checkmark & \checkmark & & 76.70   \\
\hline
\textbf{SGNet} &\checkmark & \checkmark & \checkmark & \checkmark & 77.27 \\
\hline
\end{tabular}
\caption{Performance of different settings of the proposed method. ``-A'' means ``Anchoring''.}
\label{tab:performance_ablation}
\end{table}

\paragraph{Effectiveness of VPG-Anchoring.}
To investigate the effectiveness of the proposed VPG-Anchoring, we conduct ablation experiments and introduce three different models for comparisons. The first setting is only the feature
extractor and the subnetwork of classification and regression, which is regarded as ``Base'' model. In Base, anchor is generated uniformly at all positions of the feature map, and $A_{anchor}$ is lowered to ensure the same number with SGNet. In addition, we conduct another model (``Base+V'') by adding VPG-Anchor. And we also replace the $L_{center}$ by straight line fitted directly by key points as the ``Base+V-F'' to explore the importance of VP. The comparisons of above models are listed in Tab.~\ref{tab:performance_ablation}. We can observe that the VPG-Anchoring greatly improve the performance
 of Base model, which verifies the effectiveness of this mechanism. In addition, comparing Base+V with Base+V-F, we find the proposed approximate VP in lane presentation is better than the one by direct fitting.

\paragraph{Effectiveness of Multi-level Structures.}
To explore the effectiveness of the pixel-level, lane-level and image-level structures, we conduct another experiments by combining the pixel-level perception with ``Base+V'' as ``Base+V+P'' and 
adding lane-level relation to ``Base+V+P'' as 
``Base+V+P+L''. From the last four rows of Tab.~\ref{tab:performance_ablation}, we can find that the performance of lane detection can be continuously improved by pixel-, lane- and image-level structures, which validates that the three levels of constrains are compatible with each other, and can be used together to gain performance. 

\section{Conclusion}

In this paper, we rethink the difficulties 
that hinder the development of lane detection 
and propose a structure guided framework. In this framework, we introduce a new lane representation to meet the demands of various lane representations. Based on the representation, we propose a novel vanishing point guided anchoring mechanism to generate intensive anchors for efficiently capturing lanes.  
In addition, multi-level structure constraints is modeled to improve lane perception. Extensive experiments on benchmark datasets validates the effectiveness of the proposed approach with fast inference 
and shows that the perspective of modeling and utilization of structure information is useful for lane detection.

\bibliographystyle{named}
\bibliography{SGNet}

\end{document}